\documentclass[10pt, a4paper]{article}
\usepackage[final]{lrec2026}

\usepackage{fontspec}
\usepackage{bidi}
\newfontfamily\persianfont[
  Script=Arabic,
  Path=./,
  Extension=.ttf,
  UprightFont=Amiri-Regular,
  BoldFont=Amiri-Bold,
  ItalicFont=Amiri-Italic,
  BoldItalicFont=Amiri-BoldItalic
]{Amiri}

\title{PerSoMed: A Large-Scale Balanced Dataset for Persian Social Media Text Classification}

\name{Isun Chehreh and Ebrahim Ansari} 

\address{Institute for Advanced Studies in Basic Sciences (IASBS)\\
         \{a.chehreh and ansari\}@iasbs.ac.ir\\}

\abstract{
This research introduces the first large-scale, well-balanced Persian social media text classification dataset, specifically designed to address the lack of comprehensive resources in this domain. The dataset comprises 36,000 posts across nine categories (Economic, Artistic, Sports, Political, Social, Health, Psychological, Historical, and Science \& Technology), each containing 4,000 samples to ensure balanced class distribution. Data collection involved 60,000 raw posts from various Persian social media platforms, followed by rigorous preprocessing and hybrid annotation combining ChatGPT-based few-shot prompting with human verification. To mitigate class imbalance, we employed undersampling with semantic redundancy removal and advanced data augmentation strategies integrating lexical replacement and generative prompting.
We benchmarked several models, including BiLSTM, XLM-RoBERTa (with LoRA and AdaLoRA adaptations), FaBERT, SBERT-based architectures, and the Persian-specific TookaBERT (Base and Large). Experimental results show that transformer-based models consistently outperform traditional neural networks, with TookaBERT-Large achieving the best performance (Precision: 0.9622, Recall: 0.9621, F1-
score: 0.9621). Class-wise evaluation further confirms robust performance across all categories, though social and political texts exhibited slightly lower scores due to inherent ambiguity.
This research presents a new high-quality dataset and provides comprehensive evaluations of cutting-edge models, establishing a solid foundation for further developments in Persian NLP, including trend analysis, social behavior modeling, and user classification. The dataset is publicly available to support future research endeavors.
\\ \newline
 }

\UseMicrotypeSet[protrusion]{basicmath}
\microtypesetup{nopatch=footnote}
\begin{document}

\maketitleabstract

\section{Introduction}
The rapid growth of online textual content, fueled by social media platforms, news outlets, and collaborative knowledge bases, has introduced major challenges in organizing and interpreting unstructured data (\citet{taha2024comprehensive}). Among these sources, social media posts stand out for their brevity, informality, and fast-changing nature. Such characteristics make automatic text classification—a process of assigning predefined categories to texts—both essential and difficult. This task underpins a wide range of applications, including sentiment analysis, topic identification, spam detection, and event tracking (\citet{gao2025multi}\citet{chehreh2024advanced}).

Although recent advances in deep learning and transformer-based architectures have considerably improved text classification performance (\citet{wang2024adaptable}\citet{tareh2025pabsa}), classifying short and noisy posts remains a demanding problem. These texts often lack clear linguistic structure and depend heavily on contextual cues such as user interactions or surrounding discourse.(\citet{zhang2025pushing}\citet{tareh2025iasbs}).

In this study, we address this gap by constructing a new dataset designed specifically for classification of social media posts. The dataset includes three core components—post ID, textual context, and label—covering nine distinct categories. We systematically evaluate a range of modeling strategies, from conventional neural models to transformer-based approaches, in order to assess their strengths and limitations in handling such data.

The remainder of this paper is organized as follows: Section 2 reviews related studies, emphasizing recent progress and existing challenges in text classification and domain adaptation. Section 3 outlines our methodology, with Section 3.1 describing the dataset and Section 3.2 detailing the modeling approaches. Section 4 presents the experimental findings, divided into two parts: Section 4.1 reports the main results, and Section 4.2 presents the ablation study, where we analyze the impact of the undersampling and oversampling strategies on the model’s performance. Finally, Section 5 concludes the paper with a summary of the main contributions.

\section{Related Work}

In this part of the paper, we review recent developments in text classification and Persian-language resources that are relevant to our work. The reviewed studies collectively address the challenges of handling short, noisy, and domain-specific texts, particularly those found in social media contexts.

\citet{khojasteh2020lscp} introduced LSCP, a large-scale colloquial Persian corpus comprising 120 million sentences collected from Twitter. The dataset includes multiple linguistic annotations such as POS tags, dependency trees, and sentiment polarity, and serves as a foundational resource for developing models capable of understanding informal and non-standard Persian text.

Building on this line of work, \citet{zinvandi2025famteb} presented FaMTEB, the most comprehensive benchmark to date for evaluating Persian text embedding models. FaMTEB spans 63 datasets across seven NLP tasks, including text classification, clustering, semantic textual similarity, retrieval, reranking, pair classification, and summary retrieval. The datasets are drawn from a mix of translated English corpora, curated Persian web data, and synthetic samples generated via large language models, allowing for a rich and diverse evaluation environment.

A unique strength of FaMTEB lies in its inclusion of chatbot-oriented and retrieval-augmented generation (RAG) evaluation datasets, which make it possible to assess how well embeddings support interactive and generative applications. By establishing standardized metrics and covering both general-purpose and task-specific evaluations, FaMTEB has effectively laid the groundwork for systematic benchmarking of Persian and multilingual models under low-resource conditions.

Several studies have also proposed methodological advances aimed at improving classification performance. \citet{galal2024rethinking} revisited BERT-based classifiers and proposed aggregation mechanisms that replace the standard [CLS] representation, achieving strong results on sarcasm and sentiment detection—tasks closely aligned with social media analysis. \citet{gao2025multi} enhanced transformer-based models through a multi-level attention mechanism that captures both global document structure and fine-grained contextual cues, improving robustness on short and noisy texts. Similarly, \citet{gao2024text} proposed a multimodal framework that integrates textual, visual, and metadata features, yielding notable improvements on social media datasets.

In a related line of research, \citet{chehreh2024enhanced} introduced a two-stage clustering and DeBERTa-based framework for multi-label question tagging on Stack Overflow. Their approach first clusters semantically similar questions using SMPNet and UMAP, and then fine-tunes separate DeBERTa models for each cluster. This design improves both accuracy and computational efficiency by tailoring models to homogeneous subsets of data—a strategy that is also highly relevant for handling short, domain-specific, and semantically diverse Persian social media texts.

Comparative studies of language models have provided complementary insights. \citet{bucher2024fine} and \citet{edwards2024language} showed that smaller fine-tuned transformer models (e.g., RoBERTa, DeBERTa) often outperform large generative models such as GPT-3.5 or GPT-4 in low-data scenarios, emphasizing the continued relevance of task-specific fine-tuning. \citet{wang2024adaptable} further explored model adaptability by combining prompting, few-shot learning, and fine-tuning in a unified text classification framework.

Finally, \citet{philipo2025assessing} investigated transformer-based models for cyberbullying detection, a socially important and context-dependent task. Their findings highlighted the trade-off between accuracy and computational efficiency, showing that compact architectures such as DistilBERT are well-suited for real-time, multilingual detection applications.

Overall, the reviewed studies reveal two central trends: (1) a growing focus on integrating contextual and multimodal information to better handle noisy and short texts; and (2) a strong move toward building and evaluating Persian-specific resources—such as LSCP and FaMTEB—that enable more robust and reproducible research in low-resource settings. These developments form the foundation for our study, which focuses on context-aware classification of Persian social media posts.

\section{APPROACH}
This section begins with a detailed overview of the dataset created for this investigation. We will first outline the steps involved in constructing the dataset, including data collection, pre-processing, and labeling, which form the foundation for the subsequent analysis. In the following section, we review the implemented models, covering both conventional deep learning architectures and approaches based on Transformers, to systematically evaluate their performance in our dataset and assess their suitability for handling short, noisy and context-rich social media posts.

\subsection{Dataset}
For the task of classifying Persian social media texts, no suitable datasets were available that met the requirements for effective and large-scale text classification. Existing datasets were either limited in the number of categories—typically containing only four or five classes—or suffered from small sample sizes, often around 500 samples per class, which is insufficient for developing robust classification models. This gap motivated us to construct a novel, comprehensive dataset specifically designed for Persian social media text classification.
To build this dataset, we collected approximately 60,000 posts from various Persian social media platforms. After initial preprocessing, including the removal of personally identifiable information to ensure privacy, about 50,000 posts remained for further processing.
A crucial step involved labeling these posts, for which we defined nine distinct categories to comprehensively cover the spectrum of social media content. These categories included:
\begin{itemize}
    \item Economic  ({\persianfont اقتصادی})
    \item Artistic  ({\persianfont  هنری})
    \item Sports  ({\persianfont ورزشی})
    \item Political  ({\persianfont سیاسی})
    \item Social  ({\persianfont اجتماعی})
    \item Health  ({\persianfont سلامتی})
    \item Psychological  ({\persianfont روانشناسی})
    \item Historical  ({\persianfont تاریخی})
    \item Science and Technology  ({\persianfont  فناوری و علم})
\end{itemize}

Nearly 80\% of the collected posts originated from the year 2025, reflecting contemporary trends and discourse. For labeling, the ChatGPT API (\citet{openai_api_2025}) was utilized with a few-shot prompting (\citet{ma2023fairness}). Carefully designed prompts with detailed instructions and multiple examples per category helped guide the model to accurately assign labels. Subsequently, human annotators reviewed the labeled data to remove noisy or ambiguous instances, maintaining high dataset quality.A structured dataset was created in a three-column format, where the first column represents the ID, the second column contains the context (i.e., the text of the post), and the third column specifies the label assigned to each instance.

Despite these efforts, significant class imbalance was observed. For example, the “Political” category contained 16,637 samples, whereas the “Historical” category had only 497 samples, which risked introducing bias in model training. To address this, undersampling techniques were applied to the overrepresented categories. Using ParsBERT (\citet{farahani2021parsbert}), posts were converted into semantic embeddings that capture deep contextual and syntactic relationships between Persian words and phrases. ParsBERT is a transformer-based language model pre-trained specifically for Persian text, capable of producing high-quality vector representations where semantically similar sentences are located closer in the embedding space. Leveraging these embeddings, cosine similarity (\citet{li2013distance}) was employed to identify and remove semantically redundant samples. From the remaining data, 4,000 samples with the lowest similarity scores were selected to maximize diversity.

For categories with insufficient data, synthetic data augmentation was employed. Initial attempts using back-translation(\citet{bourgeade2024data}) yielded unsatisfactory quality due to poor translation fidelity. Therefore, to effectively address the data scarcity in underrepresented categories, we employed a hybrid data augmentation strategy combining lexical replacement (\citet{coulombe2018text}) and advanced few-shot prompting using the ChatGPT API. Synonym replacement, which accounted for approximately 10\% of the augmented data, involved substituting words with their contextually appropriate synonyms to introduce lexical variation without altering the original meaning. While this method was relatively straightforward and computationally efficient, its capacity for generating diverse and semantically rich samples was limited.

To overcome these limitations, the majority of augmentation—about 90\%—relied on ChatGPT’s few-shot prompting capabilities. Carefully crafted prompts, enriched with multiple category-specific examples, guided the model to generate high-quality, coherent, and contextually relevant synthetic posts. This process, though time-consuming, yielded synthetic samples that closely mirrored the linguistic nuances and thematic characteristics of real social media texts in Persian. By iteratively refining the prompts and validating outputs, we ensured that the generated data was both diverse and representative of the targeted categories.
Overall, this combined augmentation approach significantly improved the dataset’s balance and richness, providing a robust foundation for training models that perform reliably across all classes, including those originally underrepresented.

The final balanced dataset consists of exactly 36,000 posts, with 4,000 samples per category. Our dataset is organized into three attributes: the first corresponds to the post identifier, the second represents the textual content, and the third specifies the associated label. Posts shorter than three words were excluded to maintain content richness. Statistical analysis reveals an average of 1.18 hashtags per post, with a maximum of 22 hashtags; 3,898 posts contained no hashtags, while 32,102 posts had at least one. Table~\ref{tab:frequent_hashtags} presents the ten hashtags that appear most frequently in the dataset. 

\begin{table}
\renewcommand{\arraystretch}{1.5}
\begin{tabular}{p{5.4cm}cc}
\hline
\textbf{Hashtag} & \textbf{Frequency} \\
\hline
\textbf{History ({\persianfont تاریخ})} & \textbf{1062}\\
Health ({\persianfont سلامتی}) & 734 \\
Economic ({\persianfont اقتصاد}) & 554 \\
Football ({\persianfont فوتبال}) & 424 \\
Medicine ({\persianfont پزشکی}) & 411 \\
Technology ({\persianfont فناوری}) & 352 \\
Inflation ({\persianfont تورم})& 344 \\
Health ({\persianfont سلامت}) & 334 \\
Art ({\persianfont هنر}) & 322 \\
Artificial intelligence  ({\persianfont مصنوعی هوش}) & 318 \\
\hline
\end{tabular}
\caption{\label{tab:frequent_hashtags} 10 most frequent hashtags in the dataset.}
\end{table}

The average word count per post was 17.37, ranging from 3 to 116 words. Detailed per-category statistics can be found in Table~\ref{tab:avg_word_length}.

\begin{figure}[htbp]
\begin{center}
\includegraphics[width=\columnwidth]{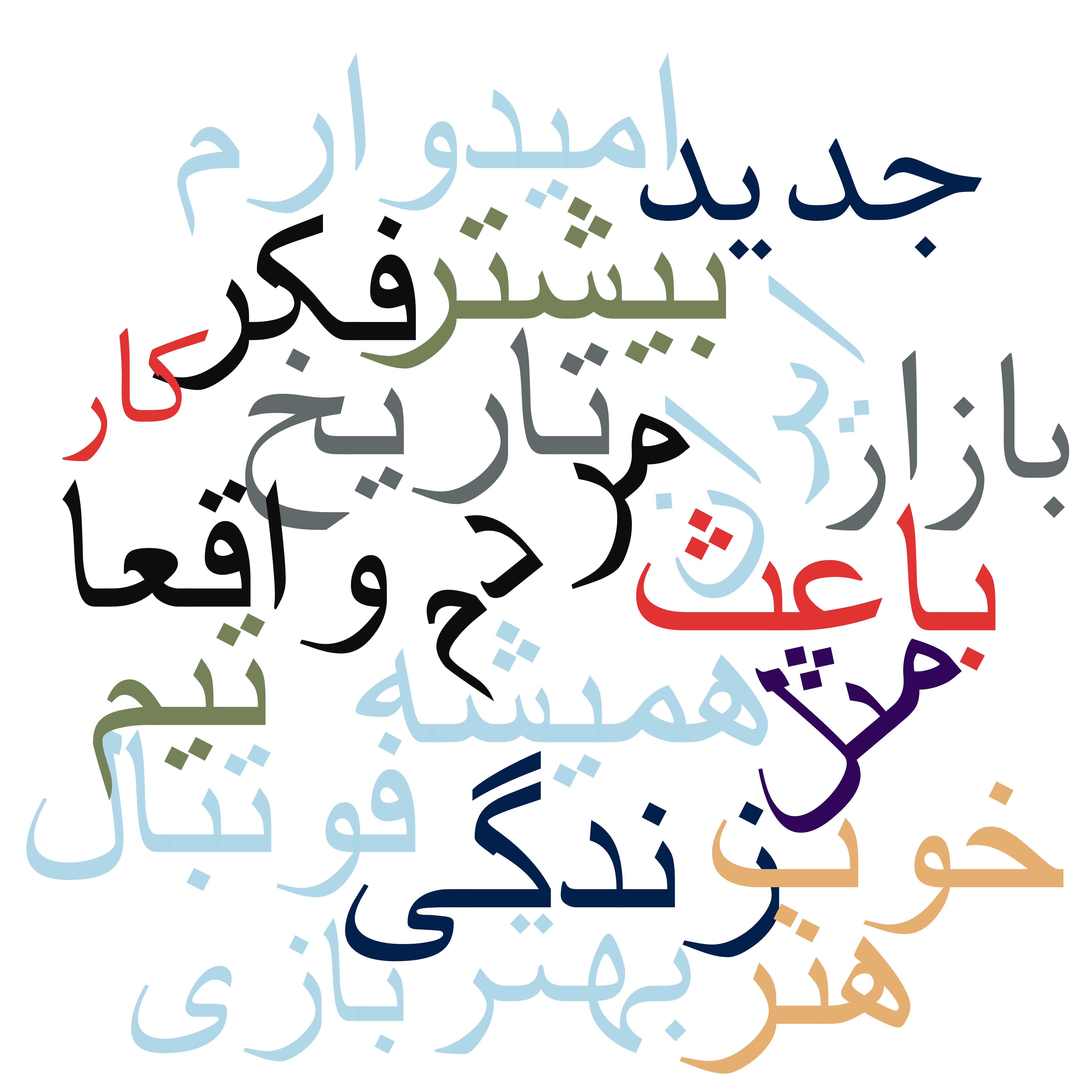}
\caption{The 20 most frequent words in the dataset.}
\label{fig:fig1}
\end{center}
\end{figure}

\begin{table}
\renewcommand{\arraystretch}{1.6}
\begin{tabular}{p{4cm}cc}
\hline
\textbf{Category} & \textbf{Average Word} \\
\hline
Social & 19.42 \\
Economic & 14.44 \\
Historical & 18.85 \\
Psychological & 16.00 \\
Health & 16.47 \\
\textbf{Political} & \textbf{20.16} \\
Science and Technology & 18.06 \\
Artistic & 17.83 \\
Sports & 15.06 \\
\hline
\end{tabular}
\caption{\label{tab:avg_word_length} Average word count per category in the dataset.}
\end{table}
After removing stopwords, the twenty most frequent words were extracted and visualized as a word cloud in Figure~\ref{fig:fig1}, with word size reflecting their frequency in the corpus.

As far as we are aware, this dataset represents the first large-scale and balanced Persian corpus designed for social media text classification across multiple categories. It offers a valuable resource for various research domains, including thematic analysis, trend detection, social behavior modeling, and notably, user classification based on shared content, specifically within Persian-language social media platforms.The complete dataset has been made publicly available and can be accessed through the \href{https://drive.google.com/file/d/1BVzeYUoDe7Rm80dK4M7ENZLPa6FYxf1V/view?usp=sharing}{Dataset Link}.

\subsection{Models}

After introducing the dataset, we apply several models to this new dataset and evaluate their performance.

The initial stage of the workflow focuses on data preprocessing, a crucial step for improving data quality and making it suitable for Natural Language Processing (NLP) models. A range of preprocessing methods was applied to clean and normalize the textual content.

A sequence of preprocessing procedures was carried out to improve the quality of the text data:

Punctuation Removal: All unnecessary punctuation marks and special characters were removed from the text. This step helps to standardize the text and prevents the negative influence of punctuation marks during the modeling process.

Text Normalization: Text normalization plays a crucial role in processing Persian texts. This step addresses writing inconsistencies such as variations in character usage (e.g., {\persianfont "ی" } vs {\persianfont "ي" }) and standardizes the text to a uniform format.

Removal of Emojis and Mentions: Emojis and social media-specific markers, such as @mentions, were eliminated from the data. These elements, while important in the context of social media, were removed to retain only the core textual content.

Elimination of Non-Persian Characters: Non-Persian characters, which occasionally appear in some texts, were identified and removed to further enhance the quality of the dataset.

Removing Extra Whitespaces: Extra spaces between words and sentences were removed, ensuring that only standard spaces remained.

These preprocessing steps are fundamental in refining the dataset and improving its suitability for NLP models. After preprocessing, the data was ready for the next stage: model training and evaluation.

After preprocessing, the cleaned dataset was used to train and evaluate several models to determine the most effective approach for Persian text classification. Both traditional neural networks and state-of-the-art transformer-based models were explored. Below, we describe each model and the methodology used for evaluation.

To establish a baseline for sequence modeling, we employed a Bidirectional Long Short-Term Memory (BiLSTM) network (\citep{graves2005framewise}). Unlike standard LSTMs, BiLSTMs read sequences in both directions, enabling the model to incorporate contextual information from preceding and succeeding tokens. This feature is particularly crucial for Persian, where sentence meaning often depends on context and word order. The architecture begins with an embedding layer that converts tokens into dense vector representations, followed by one or more BiLSTM layers to capture sequential patterns, and ends with a fully connected layer for classification. The bidirectional design allows the network to better handle complex sequences than unidirectional LSTMs or traditional recurrent structures.

One of the models employed in this study is XLM-RoBERTa Base (XLMR-B) (\citep{conneau2019unsupervised}), a large-scale multilingual transformer trained on more than 100 languages using over two terabytes of filtered CommonCrawl data. XLM-R Base utilizes the self-attention mechanism of Transformers to capture long-range dependencies and semantic relationships across multiple languages, making it a robust baseline for multilingual text classification, including tasks in Persian.
The model is trained using the masked language modeling (MLM) objective, predicting masked tokens within monolingual text sequences. Subword tokenization is performed with SentencePiece using a unigram language model. Unlike some earlier multilingual models, XLM-R Base does not use explicit language embeddings, which enhances its capability to handle code-switching. The architecture consists of 12 layers, a hidden size of 768, and approximately 270 million parameters.

To enhance computational efficiency and reduce memory demands during fine-tuning, we employed the Low-Rank Adaptation (LoRA) technique (\citep{hu2022lora}) on XLM-RoBERTa. LoRA is a parameter-efficient method that integrates trainable low-rank matrices into transformer layers while keeping the original model weights fixed. This approach enables fine-tuning on specific tasks using substantially fewer trainable parameters, thereby lowering both memory consumption and training time, while achieving performance comparable to—or even exceeding—full fine-tuning. In our study, LoRA was applied specifically to the attention layers of XLM-RoBERTa.

In addition to the standard LoRA approach, we implemented an adaptive variant, XLM-RoBERTa Base with AdaLoRA (\citep{zhang2023adalora}), tailored for Persian text classification. AdaLoRA dynamically distributes low-rank matrices across transformer layers based on the relative significance of each layer. By allocating more capacity to critical layers while reducing resources for less important ones, this approach enables more efficient and flexible fine-tuning. The design decreases the number of trainable parameters and computational demands without compromising performance, making it especially well-suited for large pre-trained models in resource-limited environments.

We also utilized FaBERT (\citep{masumi2024fabert}), a Persian BERT-based model pre-trained on the HmBlogs corpus, which encompasses both formal and informal Persian texts. FaBERT is tailored to capture the unique linguistic characteristics of Persian, including its morphology, syntax, and flexible sentence structures, which are crucial for precise text classification. By leveraging language-specific pretraining, the model effectively manages both standard and colloquial expressions, demonstrating enhanced performance on downstream tasks such as sentiment analysis, question answering, and natural language inference. Its compact design supports efficient training and inference while maintaining high accuracy, making it particularly suitable for environments with limited computational resources.

We also utilized Paraphrase-Multilingual-MiniLM-L12-v2 (\citep{korga2025does}), a multilingual sentence embedding model from the Sentence-BERT (SBERT) family (\citep{reimers2019sentence}), designed specifically for tasks such as semantic similarity and paraphrase detection. Its ability to process multiple languages makes it suitable for cross-lingual applications, including job recommendation systems.

To improve the model’s capacity to capture complex semantic relationships, we augmented the base model with multi-head self-attention layers (\citep{voita2019analyzing}), followed by fully connected layers for classification. These additional attention layers allow the model to learn detailed inter-token dependencies that may not be fully captured by the base embeddings. Studies on multi-head self-attention indicate that only a subset of attention heads is critical for performance, often corresponding to interpretable linguistic functions such as capturing positional or syntactic information, while the remaining heads can be pruned without significantly affecting accuracy. This architecture therefore provides a balance between expressiveness and computational efficiency.

The model comprises 12 transformer layers, following the MiniLM design, with a total of 118 million parameters. Fine-tuning was performed using a triplet loss approach, which encourages semantically similar sentences to be embedded closer together while separating dissimilar ones. This strategy enhances the model’s ability to detect subtle semantic similarities across varied text inputs.

By combining pretrained multilingual embeddings, multi-head self-attention, and fully connected classification layers, this approach achieves an effective balance between accuracy and efficiency, making it well-suited for large-scale semantic similarity tasks, cross-lingual text comparison, and recommendation systems.

The next models considered in this study are TookaBERT-Base and TookaBERT-Large (\citep{sadraeijavaheri2024tookabert}), both Transformer-based language models pre-trained on extensive Persian corpora. TookaBERT-Base follows the BERT-base architecture, comprising 12 transformer layers with multi-head self-attention and feed-forward networks. It employs Byte-Pair Encoding (BPE) for tokenization and whole-word masking during pre-training, enabling the model to effectively capture Persian’s rich morphological and syntactic structures.

To enhance training efficiency, Flash Attention v2 and the ZeRO Stage 2 optimizer are applied, promoting faster convergence and efficient GPU memory utilization. This setup provides a balanced compromise between computational cost and performance, making TookaBERT-Base well-suited for resource-constrained environments while maintaining strong results on Persian natural language understanding (NLU) tasks.

TookaBERT-Large builds upon the Base architecture but significantly increases both depth and capacity. It contains 24 transformer layers and more attention heads per layer, allowing it to capture more complex semantic patterns and fine-grained linguistic nuances in Persian texts. As a result, the Large version delivers superior performance across various NLU tasks, including question answering, sentiment analysis, and natural language inference. However, this increased representational power comes with higher computational and memory requirements.

For evaluation, standard metrics including F1-score, Precision, and Recall were employed to ensure a fair and consistent comparison across all models. These metrics provide complementary perspectives: Precision measures the accuracy of positive predictions, Recall assesses the ability to capture relevant instances, and F1-score balances both to give an overall view of performance. 

\section{Experiments}  
This section presents the experimental findings and is divided into two parts. First, we report the main results, highlighting the performance of models on the preprocessed dataset. Second, we present the ablation study, analyzing the impact of undersampling and oversampling.

\subsection{Results}

\begin{table}[h!]
\centering  
\renewcommand{\arraystretch}{3} 
\begin{tabular}{m{2cm}>
{\centering\arraybackslash}m{1.45cm}>{\centering\arraybackslash}m{1.45cm}>{\centering\arraybackslash}m{1.45cm}}
\hline
\textbf{Model} & \textbf{Precision} & \textbf{Recall} & \textbf{F1} \\
\hline
BiLSTM & 0.8823 & 0.8807 & 0.8813 \\
\hline
SBERT     +Attention       +FCLayers & \centering{0.8950} & 0.8760 & 0.8850 \\
\hline
FaBERT & 0.9285 & 0.9272 & 0.9269 \\
\hline
XLMR-B & 0.9494 & 0.9490 & 0.9491 \\
\hline
XLMR-B +LoRA & 0.9401 & 0.9407 & 0.9403 \\
\hline
XLMR-B +AdaLoRA & 0.9237 & 0.9246 & 0.9239 \\
\hline
TookaBERT\-Base & 0.9354 & 0.9358 & 0.9354 \\
\hline
\textbf{TookaBERT\-Large} & \textbf{0.9622} & \textbf{0.9621} & \textbf{0.9621} \\
\hline
\end{tabular}
\caption{\label{tab:model_results} Performance evaluation of different models for Persian text classification.}
\end{table}

The models were trained and assessed using the preprocessed dataset. Table \ref{tab:model_results} presents a summary of their performance across standard evaluation metrics, including precision, recall, and F1-score. The findings show that transformer-based models generally surpass the BiLSTM baseline. Among these, TookaBERT-Large achieved the best results, highlighting the benefits of language-specific pretraining for Persian text. XLM-RoBERTa variants, such as LoRA and AdaLoRA, also delivered strong performance while providing parameter-efficient alternatives. FaBERT demonstrated solid results, emphasizing the value of Persian-focused language representations. Additionally, enhancing the multilingual MiniLM embeddings with multi-head self-attention yielded moderate improvements compared to simpler baseline models.

Overall, the evaluation shows that leveraging transformer-based models with either multilingual or Persian-specific pretraining significantly improves the classification of social media text.

Table \ref{tab:model_results} summarizes the overall performance of all implemented models. Among them, TookaBERT-Large achieved the best results, demonstrating the effectiveness of Persian-specific pretraining.

To further analyze its performance, a class-wise evaluation was conducted, as shown in Table \ref{tab:classwise_results}. The results indicate that TookaBERT-Large performs consistently well across all categories. While most classes, such as Historical, Psychological, and Sports, achieved near-perfect scores, the Social and Political categories showed slightly lower performance. This discrepancy may be due to the inherent complexity and ambiguity of social and political texts in social media. Overall, the high precision, recall, and F1-scores across all classes confirm that TookaBERT-Large is highly effective for Persian social media text classification.

\begin{table}[h!]
\centering
\renewcommand{\arraystretch}{1.9} 
\begin{tabular}{m{2cm}>
{\centering\arraybackslash}m{1.4cm}>{\centering\arraybackslash}m{1.4cm}>{\centering\arraybackslash}m{1.4cm}}
\hline
\textbf{Category} & \textbf{Precision} & \textbf{Recall} & \textbf{F1} \\
\hline
Social & 0.88 & 0.88 & 0.88\\
Economic & 0.98 & 0.98 & 0.98\\
Historical & 1.00 & 0.99 & 0.99\\
Psychological & 0.98 & 0.98 & 0.98\\
Health & 0.97 & 0.96 & 0.96 \\
Political & 0.91 & 0.93 & 0.92 \\
Science and Technology & 0.98 & 0.98 & 0.98 \\
Artistic & 0.98 & 0.97 & 0.97\\
Sports & 0.99 & 0.99 & 0.99\\
\hline
\end{tabular}
\caption{\label{tab:classwise_results} Class-wise performance of TookaBERT-Large.}
\end{table}

\subsection{Ablation Study}
To evaluate the impact of dataset balancing techniques, we trained and tested several models on the original, unbalanced dataset described in the previous sections. This allowed us to observe their baseline performance before applying any undersampling or oversampling strategies. The results are presented in Table~\ref{tab:models}. As expected, all models perform worse on the unbalanced data. BiLSTM, for example, reached an F1-score of 0.3394, while TookaBERT-Large achieved 0.8621, both lower than their performance on the balanced dataset. Other models, including XLM-RoBERTa variants and FaBERT, also showed decreased scores, confirming that class imbalance negatively affects model performance, even for strong transformer-based architectures.

Notably, the BiLSTM model exhibited the largest performance drop among all models when trained on unbalanced data. However, after applying balancing techniques, it also showed the most significant improvement, indicating its high sensitivity to class imbalance. This suggests that traditional neural architectures like BiLSTM are more vulnerable to skewed class distributions compared to transformer-based models, but they can also benefit substantially from proper data balancing strategies.

Applying undersampling and oversampling significantly improves results. By reducing redundancy in overrepresented classes and augmenting underrepresented ones, models can achieve more consistent performance across all categories. In particular, Persian-specific models like TookaBERT-Large benefit the most, reaching an F1-score of 0.9621 after balancing, which demonstrates the effectiveness of these techniques.

Overall, this ablation study highlights that addressing class imbalance is crucial for Persian social media text classification. While transformer-based models already provide a strong foundation, proper data balancing ensures reliable and uniform performance across all categories.

\begin{table}[ht!]
\centering
\renewcommand{\arraystretch}{2.2} 
\begin{tabular}{m{2cm}>
{\centering\arraybackslash}m{1.4cm}>{\centering\arraybackslash}m{1.4cm}>{\centering\arraybackslash}m{1.4cm}}
\hline
\textbf{Model} & \textbf{Recall} & \textbf{Precision} & \textbf{F1} \\
\hline
BiLSTM & 0.3127 & 0.3703 & 0.3394 \\
XLMR-B & 0.6747 & 0.6700 & 0.6725 \\
XLMR-B +LORA & 0.7390 & 0.7560 & 0.7473 \\
Fabert & 0.7697 & 0.7639 & 0.7668 \\
TookaBERT-Base & 0.7514 & 0.7567 & 0.7538 \\
\textbf{TookaBERT-Large} & \textbf{0.8621} & \textbf{0.8622} & \textbf{0.8621} \\

\hline
\end{tabular}
\caption{\label{tab:models}Evaluation results of several models on the unbalanced dataset, before applying undersampling and oversampling.}
\end{table}

\section{Conclusion}
In this study, we introduced the first large-scale, balanced dataset for Persian social media text classification, addressing a critical gap in the availability of high-quality resources for Persian-language NLP. By collecting, preprocessing, and carefully annotating 60,000 posts across nine diverse categories, we constructed a robust dataset of 36,000 samples with equal class distribution. The hybrid data augmentation strategy—combining lexical replacement with advanced few-shot prompting using the ChatGPT API—proved effective in mitigating data imbalance while maintaining linguistic richness and contextual authenticity.

To evaluate the dataset, we explored a wide spectrum of models ranging from BiLSTM baselines to cutting-edge transformer architectures. Our results demonstrated the clear superiority of transformer-based models, with TookaBERT-Large achieving state-of-the-art performance. Additionally, parameter-efficient techniques such as LoRA and AdaLoRA showcased promising trade-offs between performance and computational cost, highlighting their potential for deployment in resource-constrained environments.

Beyond providing a benchmark dataset and empirical comparisons, this research offers several key contributions: (1) establishing a standardized resource for Persian social media analysis, (2) demonstrating the effectiveness of hybrid augmentation strategies for low-resource languages.

Despite these advances, certain challenges remain. The subtle overlap and inherent ambiguity in categories such as “Social” and “Political” limit classification accuracy, suggesting that future work should explore hierarchical or multi-label classification approaches.

Overall, this work lays the foundation for advancing Persian NLP research in social media contexts, enabling applications such as trend detection, user profiling, misinformation analysis, and sociolinguistic studies. We hope that both the released dataset and our comparative analysis of state-of-the-art models will inspire further research and foster the development of more inclusive NLP tools for Persian.

\section{Ethics Statement}
This work involves the collection of publicly available Persian social media posts. No personally identifiable information was stored or published. Usernames, profile information, and URLs were removed during preprocessing. All data was processed in accordance with ethical research guidelines for handling public online content.

\section{Limitations}
While the dataset covers nine major categories, the boundaries between some classes (e.g., “Social” and “Political”) can be ambiguous, leading to lower classification scores in these areas. Moreover, the dataset reflects recent online discourse and may not fully generalize to future linguistic or topical trends.

In addition, due to the limited number of collected samples and the inability to cover all time periods, the dataset may reflect specific socio-political contexts or events, such as presidential elections or ongoing conflicts.

\section{Bibliographical References}\label{sec:reference}
\bibliographystyle{lrec2026-natbib}

\begin{thebibliography}{30}
\expandafter\ifx\csname natexlab\endcsname\relax\def\natexlab#1{#1}\fi

\bibitem[{Bourgeade et~al.(2024)Bourgeade, Casola, Wizani, Bosco et~al.}]{bourgeade2024data}
Tom Bourgeade, Silvia Casola, Adel~Mahmoud Wizani, Cristina Bosco, et~al. 2024.
\newblock Data augmentation through back-translation for stereotypes and irony detection.
\newblock In \emph{CEUR WORKSHOP PROCEEDINGS}, volume 3878, pages 90--97. CEUR-WS.

\bibitem[{Bucher and Martini(2024)}]{bucher2024fine}
Martin Juan~Jos{\'e} Bucher and Marco Martini. 2024.
\newblock Fine-tuned'small'llms (still) significantly outperform zero-shot generative ai models in text classification.
\newblock \emph{arXiv preprint arXiv:2406.08660}.

\bibitem[{Chehreh et~al.(2024{\natexlab{a}})Chehreh, Ansari, and Bigham}]{chehreh2024advanced}
Isun Chehreh, Ebrahim Ansari, and Bahram~Sadeghi Bigham. 2024{\natexlab{a}}.
\newblock Advanced automated tagging for stack overflow: A multi-stage approach using deep learning and nlp techniques.
\newblock In \emph{2024 20th CSI International Symposium on Artificial Intelligence and Signal Processing (AISP)}, pages 1--6. IEEE.

\bibitem[{Chehreh et~al.(2024{\natexlab{b}})Chehreh, Saadati, Ansari, and Bigham}]{chehreh2024enhanced}
Isun Chehreh, Farzaneh Saadati, Ebrahim Ansari, and Bahram~Sadeghi Bigham. 2024{\natexlab{b}}.
\newblock Enhanced multi-label question tagging on stack overflow: A two-stage clustering and deberta-based approach.
\newblock In \emph{Proceedings of the 36th Conference of Open Innovations Association FRUCT}.

\bibitem[{Conneau et~al.(2019)Conneau, Khandelwal, Goyal, Chaudhary, Wenzek, Guzm{\'a}n, Grave, Ott, Zettlemoyer, and Stoyanov}]{conneau2019unsupervised}
Alexis Conneau, Kartikay Khandelwal, Naman Goyal, Vishrav Chaudhary, Guillaume Wenzek, Francisco Guzm{\'a}n, Edouard Grave, Myle Ott, Luke Zettlemoyer, and Veselin Stoyanov. 2019.
\newblock Unsupervised cross-lingual representation learning at scale.
\newblock \emph{arXiv preprint arXiv:1911.02116}.

\bibitem[{Coulombe(2018)}]{coulombe2018text}
Claude Coulombe. 2018.
\newblock Text data augmentation made simple by leveraging nlp cloud apis.
\newblock \emph{arXiv preprint arXiv:1812.04718}.

\bibitem[{Edwards and Camacho-Collados(2024)}]{edwards2024language}
Aleksandra Edwards and Jose Camacho-Collados. 2024.
\newblock Language models for text classification: Is in-context learning enough?
\newblock \emph{arXiv preprint arXiv:2403.17661}.

\bibitem[{Farahani et~al.(2021)Farahani, Gharachorloo, Farahani, and Manthouri}]{farahani2021parsbert}
Mehrdad Farahani, Mohammad Gharachorloo, Marzieh Farahani, and Mohammad Manthouri. 2021.
\newblock Parsbert: Transformer-based model for persian language understanding.
\newblock \emph{Neural Processing Letters}, 53(6):3831--3847.

\bibitem[{Galal et~al.(2024)Galal, Abdel-Gawad, and Farouk}]{galal2024rethinking}
Omar Galal, Ahmed~H Abdel-Gawad, and Mona Farouk. 2024.
\newblock Rethinking of bert sentence embedding for text classification.
\newblock \emph{Neural Computing and Applications}, 36(32):20245--20258.

\bibitem[{Gao et~al.(2024)Gao, Yang, Sun, Xia, Ma, and Zhu}]{gao2024text}
Erdi Gao, Haowei Yang, Dan Sun, Haohao Xia, Yuhan Ma, and Yuanjing Zhu. 2024.
\newblock Text classification optimization algorithm based on graph neural network.
\newblock In \emph{2024 IEEE 6th International Conference on Power, Intelligent Computing and Systems (ICPICS)}, pages 814--822. IEEE.

\bibitem[{Gao et~al.(2025)Gao, Liu, Zhu, Zhou, Zheng, and Liao}]{gao2025multi}
IstJia Gao, Guiran Liu, Binrong Zhu, Shicheng Zhou, Hongye Zheng, and Xiaoxuan Liao. 2025.
\newblock Multi-level attention and contrastive learning for enhanced text classification with an optimized transformer.
\newblock In \emph{2025 5th International Conference on Consumer Electronics and Computer Engineering (ICCECE)}, pages 499--503. IEEE.

\bibitem[{Graves and Schmidhuber(2005)}]{graves2005framewise}
Alex Graves and J{\"u}rgen Schmidhuber. 2005.
\newblock Framewise phoneme classification with bidirectional lstm networks.
\newblock In \emph{Proceedings. 2005 IEEE International Joint Conference on Neural Networks, 2005.}, volume~4, pages 2047--2052. IEEE.

\bibitem[{Hu et~al.(2022)Hu, Shen, Wallis, Allen-Zhu, Li, Wang, Wang, Chen et~al.}]{hu2022lora}
Edward~J Hu, Yelong Shen, Phillip Wallis, Zeyuan Allen-Zhu, Yuanzhi Li, Shean Wang, Lu~Wang, Weizhu Chen, et~al. 2022.
\newblock Lora: Low-rank adaptation of large language models.
\newblock \emph{ICLR}, 1(2):3.

\bibitem[{Khojasteh et~al.(2020)Khojasteh, Ansari, and Bohlouli}]{khojasteh2020lscp}
Hadi~Abdi Khojasteh, Ebrahim Ansari, and Mahdi Bohlouli. 2020.
\newblock Lscp: Enhanced large scale colloquial persian language understanding.
\newblock \emph{arXiv preprint arXiv:2003.06499}.

\bibitem[{Korga et~al.(2025)Korga, Wefers, Hanken, Tareaf, Steemers, and Avvad}]{korga2025does}
Alex~Maximilian Korga, Sebastian Wefers, Keno Hanken, Raad~Bin Tareaf, Ben Steemers, and Hunaida Avvad. 2025.
\newblock Does size matter? examining sentence similarity performance in large language models.
\newblock In \emph{2025 International Conference on Information Networking (ICOIN)}, pages 595--600. IEEE.

\bibitem[{Li and Han(2013)}]{li2013distance}
Baoli Li and Liping Han. 2013.
\newblock Distance weighted cosine similarity measure for text classification.
\newblock In \emph{International conference on intelligent data engineering and automated learning}, pages 611--618. Springer.

\bibitem[{Ma et~al.(2023)Ma, Zhang, Bian, Liu, Zhang, Zhao, Zhang, Fu, Hu, and Wu}]{ma2023fairness}
Huan Ma, Changqing Zhang, Yatao Bian, Lemao Liu, Zhirui Zhang, Peilin Zhao, Shu Zhang, Huazhu Fu, Qinghua Hu, and Bingzhe Wu. 2023.
\newblock Fairness-guided few-shot prompting for large language models.
\newblock \emph{Advances in Neural Information Processing Systems}, 36:43136--43155.

\bibitem[{Masumi et~al.(2024)Masumi, Majd, Shamsfard, and Beigy}]{masumi2024fabert}
Mostafa Masumi, Seyed~Soroush Majd, Mehrnoush Shamsfard, and Hamid Beigy. 2024.
\newblock Fabert: Pre-training bert on persian blogs.
\newblock \emph{arXiv preprint arXiv:2402.06617}.

\bibitem[{OpenAI(2025)}]{openai_api_2025}
OpenAI. 2025.
\newblock Api reference.
\newblock \url{https://platform.openai.com/docs/api-reference}.
\newblock Accessed: 2025-08-12.

\bibitem[{Philipo et~al.(2025)Philipo, Sarwatt, Ding, Daneshmand, and Ning}]{philipo2025assessing}
Adamu~Gaston Philipo, Doreen~Sebastian Sarwatt, Jianguo Ding, Mahmoud Daneshmand, and Huansheng Ning. 2025.
\newblock Assessing text classification methods for cyberbullying detection on social media platforms.
\newblock \emph{IEEE Transactions on Information Forensics and Security}.

\bibitem[{Reimers and Gurevych(2019)}]{reimers2019sentence}
Nils Reimers and Iryna Gurevych. 2019.
\newblock Sentence-bert: Sentence embeddings using siamese bert-networks.
\newblock \emph{arXiv preprint arXiv:1908.10084}.

\bibitem[{SadraeiJavaheri et~al.(2024)SadraeiJavaheri, Moghaddaszadeh, Molazadeh, Naeiji, Aghababaloo, Rafiee, Amirmahani, Abedini, Sheikhi, and Salehoof}]{sadraeijavaheri2024tookabert}
MohammadAli SadraeiJavaheri, Ali Moghaddaszadeh, Milad Molazadeh, Fariba Naeiji, Farnaz Aghababaloo, Hamideh Rafiee, Zahra Amirmahani, Tohid Abedini, Fatemeh~Zahra Sheikhi, and Amirmohammad Salehoof. 2024.
\newblock Tookabert: A step forward for persian nlu.
\newblock \emph{arXiv preprint arXiv:2407.16382}.

\bibitem[{Taha et~al.(2024)Taha, Yoo, Yeun, Homouz, and Taha}]{taha2024comprehensive}
Kamal Taha, Paul~D Yoo, Chan Yeun, Dirar Homouz, and Aya Taha. 2024.
\newblock A comprehensive survey of text classification techniques and their research applications: Observational and experimental insights.
\newblock \emph{Computer Science Review}, 54:100664.

\bibitem[{Tareh et~al.(2025{\natexlab{a}})Tareh, Mohammadzadeh, Mohandesi, and Ansari}]{tareh2025iasbs}
Mehrzad Tareh, Erfan Mohammadzadeh, Aydin Mohandesi, and Ebrahim Ansari. 2025{\natexlab{a}}.
\newblock Iasbs at semeval-2025 task 11: Ensembling transformers for bridging the gap in text-based emotion detection.
\newblock In \emph{Proceedings of the 19th International Workshop on Semantic Evaluation (SemEval-2025)}, pages 695--702.

\bibitem[{Tareh et~al.(2025{\natexlab{b}})Tareh, Mohandesi, and Ansari}]{tareh2025pabsa}
Mehrzad Tareh, Aydin Mohandesi, and Ebrahim Ansari. 2025{\natexlab{b}}.
\newblock Pabsa: Hybrid framework for persian aspect-based sentiment analysis.
\newblock \emph{arXiv preprint arXiv:2510.04291}.

\bibitem[{Voita et~al.(2019)Voita, Talbot, Moiseev, Sennrich, and Titov}]{voita2019analyzing}
Elena Voita, David Talbot, Fedor Moiseev, Rico Sennrich, and Ivan Titov. 2019.
\newblock Analyzing multi-head self-attention: Specialized heads do the heavy lifting, the rest can be pruned.
\newblock \emph{arXiv preprint arXiv:1905.09418}.

\bibitem[{Wang et~al.(2024)Wang, Pang, Lin, and Zhu}]{wang2024adaptable}
Zhiqiang Wang, Yiran Pang, Yanbin Lin, and Xingquan Zhu. 2024.
\newblock Adaptable and reliable text classification using large language models.
\newblock In \emph{2024 IEEE International Conference on Data Mining Workshops (ICDMW)}, pages 67--74. IEEE.

\bibitem[{Zhang et~al.(2023)Zhang, Chen, Bukharin, Karampatziakis, He, Cheng, Chen, and Zhao}]{zhang2023adalora}
Qingru Zhang, Minshuo Chen, Alexander Bukharin, Nikos Karampatziakis, Pengcheng He, Yu~Cheng, Weizhu Chen, and Tuo Zhao. 2023.
\newblock Adalora: Adaptive budget allocation for parameter-efficient fine-tuning.
\newblock \emph{arXiv preprint arXiv:2303.10512}.

\bibitem[{Zhang et~al.(2025)Zhang, Wang, Li, Tiwari, and Qin}]{zhang2025pushing}
Yazhou Zhang, Mengyao Wang, Qiuchi Li, Prayag Tiwari, and Jing Qin. 2025.
\newblock Pushing the limit of llm capacity for text classification.
\newblock In \emph{Companion Proceedings of the ACM on Web Conference 2025}, pages 1524--1528.

\bibitem[{Zinvandi et~al.(2025)Zinvandi, Alikhani, Sarmadi, Pourbahman, Arvin, Kazemi, and Amini}]{zinvandi2025famteb}
Erfan Zinvandi, Morteza Alikhani, Mehran Sarmadi, Zahra Pourbahman, Sepehr Arvin, Reza Kazemi, and Arash Amini. 2025.
\newblock Famteb: Massive text embedding benchmark in persian language.
\newblock \emph{arXiv preprint arXiv:2502.11571}.

\end{thebibliography}


\end{document}